\documentclass{article}
\usepackage{arxiv}

\usepackage{graphicx}%
\usepackage{multirow}%
\usepackage{amsmath,amssymb,amsfonts}%
\usepackage{amsthm}%
\usepackage{mathrsfs}%
\usepackage[title]{appendix}%
\usepackage{xcolor}%
\usepackage{textcomp}%
\usepackage{manyfoot}%
\usepackage{booktabs}%
\usepackage{algorithm}%
\usepackage{algorithmicx}%
\usepackage{algpseudocode}%
\usepackage{listings}%
\usepackage{natbib}

\usepackage[utf8]{inputenc} % allow utf-8 input
\usepackage[T1]{fontenc}    % use 8-bit T1 fonts
\usepackage{hyperref}       % hyperlinks
\usepackage{url}            % simple URL typesetting
\usepackage{booktabs}       % professional-quality tables
\usepackage{amsfonts}       % blackboard math symbols
\usepackage{nicefrac}       % compact symbols for 1/2, etc.
\usepackage{microtype}      % microtypography
\usepackage{xcolor}         % colors
\usepackage{enumitem}

\usepackage{bbm}
\usepackage{algorithm}
\usepackage{subcaption}
\DeclareMathOperator{\EX}{\mathbb{E}}
\newcommand{\argmax}{\mathop{\mathrm{argmax}}}

\title{Beyond Random Noise: Insights on Anonymization Strategies from a Latent Bandit Study}

\author{ \href{https://orcid.org/0000-0002-7453-9186}{\includegraphics[scale=0.06]{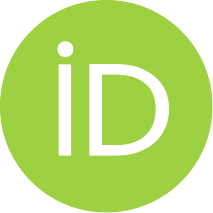}\hspace{1mm}Alexander Galozy}\thanks{Use footnote for providing further
		information about author (webpage, alternative
		address)---\emph{not} for acknowledging funding agencies.} \\
	Center for Applied Intelligent Systems Research (CAISR)\
	Halmstad University\\
	Halmstad , 30118\\
	\texttt{alexander.galozy@hh.se} \\
	\And
	\href{https://orcid.org/0000-0002-5380-4358}{\includegraphics[scale=0.06]{orcid.pdf}\hspace{1mm}Sadi Alawadi} \\
	Department of Computer Science (DIDA)\\
	Blekinge Institute of Technology\\
	Karlskrona, 37179 \\
	\texttt{sadi.alawadi@bth.se} \\
    \And
	\href{https://orcid.org/0000-0003-4071-4596}{\includegraphics[scale=0.06]{orcid.pdf}\hspace{1mm}Victor Kebande} \\
	Department of Computer Science (DIDA)\\
	Blekinge Institute of Technology\\
	Karlskrona, 37179 \\
	\texttt{victor.kebande@bth.se} \\
     \And
     \href{https://orcid.org/0000-0002-7796-5201}{\includegraphics[scale=0.06]{orcid.pdf}\hspace{1mm}S\l{}awomir Nowaczyk} \\
      Center for Applied Intelligent Systems Research (CAISR)\
	   Halmstad University\\
	   Halmstad , 30118\\
    	\texttt{slawomir.nowaczyk@hh.se} \\
}
\begin{document}

\maketitle

\begin{abstract}

This paper investigates the issue of privacy in a  learning scenario where users share knowledge for a recommendation task. Our study contributes to the growing body of research on privacy-preserving machine learning and underscores the need for tailored privacy techniques that address specific attack patterns rather than relying on one-size-fits-all solutions. We use the latent bandit setting to evaluate the trade-off between privacy and recommender performance by employing various aggregation strategies, such as averaging, nearest neighbor, and clustering combined with noise injection. More specifically, we simulate a linkage attack scenario leveraging publicly available auxiliary information acquired by the adversary.
Our results on three open real-world datasets reveal that adding noise using the Laplace mechanism to an individual user's data record is a poor choice. It provides the highest regret for any noise level, relative to de-anonymization probability and the ADS metric. 
Instead, one should combine noise with appropriate aggregation strategies. For example, using averages from clusters of different sizes provides flexibility not achievable by varying the amount of noise alone. Generally, no single aggregation strategy can consistently achieve the optimum regret for a given desired level of privacy.

%Global averaging generally provides good privacy levels but comes with significant regret. The nearest-neighbor approach and clustering provide a more favorable balance between the two metrics. 

\end{abstract}

\keywords{Latent-bandit, Privacy, Linkage-attack}

\section{Introduction}
In the era of data-driven technologies and applications, privacy-preserving machine learning has emerged as a critical concern, for example, in recommendation-based tasks which heavily rely on sensitive data. Essential steps should be considered to protect such data, especially with the introduction of stricter privacy policies like GDPR (General Data Protection Regulation)~\citep{gdpr} in Europe, which further emphasizes the necessity for enhanced privacy measures. Given the vulnerability of traditional anonymization techniques to adversarial attacks, addressing the risks associated with inadvertently disclosing sensitive data is imperative.

% Traditional anonymization techniques are vulnerable to adversarial attacks, making addressing the risk of inadvertently disclosing sensitive data imperative.

% In the current era of data-driven technologies and applications, privacy-preservation machine learning has become a critical concern, which has seen extensive utilization of information in recommendation-based tasks. As a result, the demand for personalized recommendations has seen exponential growth. However, there is a need to protect sensitive data, especially with the introduction of GDPR~\cite{gdpr}, owing to the susceptibility of traditional anonymization techniques to adversaries and attacks, and a growing concern where a machine learning model may release sensitive information about training data. 

%maybe already talk here about what this paper is about, One sentence or two is enough. It gives the reader an idea very early on in the paper and hopefully keep his/her focus :-) I will dump some text then we can slice what is not useful out

% To provide a guarantee of provable privacy, Differential Privacy (DP) has been accepted widely as a basic metric for measuring privacy 

To mitigate the problem, Differential Privacy (DP) has gained widespread acceptance as a fundamental metric for quantifying and ensuring provable privacy~\citep{dwork2014algorithmic}. However, DP has failed, for example, in federated learning setting~\citep{wang2021protecting}. Other strategies, such as homomorphic encryption, come with increased costs of computation~\citep{hao2019towards}. % To address all these limitations, alternative strategies focused on privacy protection with computation efficiency need to be explored.
Therefore, this study investigates privacy-preserving machine-learning aspects and provides valuable insights into anonymization strategies beyond the conventional use of noise, striking a balance between privacy protection and recommender system performance, exemplified in the context of latent bandits.

To assess the privacy implication of our work, we investigate the potential weaknesses associated with a linkage attack scenario. Our study assumes a susceptibility to privacy breaches due to knowledge sharing among agents. If not adequately mitigated, this weakness can enable potential adversaries to re-identify anonymized records in data via collected auxiliary information.
%
%ALTERNATIVE TEXT OF THE PREVIOUS PARAGRAPH
%To evaluate the impact on privacy of our work, we are examining the vulnerabilities related to a linkage attack scenario. Our investigation focuses on identifying weaknesses that could lead to privacy breaches exemplified using a latent bandit setting. If these weaknesses are not properly addressed, they could enable adversaries to uncover patterns and correlations in data, potentially allowing inference of private information.
%
%To assess the security implications of our work, we are basing our investigation on assumptions that portray the susceptibility, vulnerability, and prevalence of attacks in latent bandits. Specifically, we assume that an attacker has access to auxiliary information based on the system's recommendation sequences \citet{wang2020global}, that can be used to compromise the transition matrix based on the following: While differential privacy could offer a unique approach to the transition matrix by adding random noise to protect data points%by securing the statistical database\citet{vasa2022deep}%
%
%, an attacker can obtain some cells of a transition matrix through outside sources, which ultimately can allow inference of other related cells based on the nearest neighbor where a close match in the data point is identified or through statistical analysis such as clustering. On the same note, such a vulnerability could allow an attacker to employ a cluster attack to estimate the original values based on the cluster average. 
%
%\setlist{leftmargin=+12pt,itemsep=0.1em,topsep=-0.7\parskip}
The contributions of this paper can be summarised as follows:
\begin{itemize}

    \item We demonstrate the susceptibility to attacks of simple additive noise strategies and investigate alternatives with better privacy-performance trade-offs.

    \item We conduct our experiments using real-world datasets such as CASAS, Endomondo, and Fitbit, highlighting the importance of implementing tailored privatization strategies for each dataset.
    
    % We experiment on real-world CASAS, Endomundo, and Fitbit datasets showing the need for privatization strategies tailored to individual datasets.

    \item We are the first to explore privacy concerns within a learning setup in the latent bandit context, specifically in a realistic and frequently encountered linkage attack scenario.
    
    % To the best of our knowledge, we are the first to investigate privacy in a learning setup using the latent bandit setting under a realistic and commonly occurring linkage-attack scenario.
    
    \item We investigate the popular ADS metric for estimating a dataset's privacy level, demonstrating a significant mismatch between the metric and the level of de-anonymization in linkage attacks.  
\end{itemize}

%linkage attack scenario 

\section{Related Work}

The field of privacy-preserving machine learning has witnessed significant advancements in recent years. Techniques such as differential privacy (DP) ~\citep{dwork2014algorithmic} and federated learning (FL)~\citep{mcmahan17a} have emerged as prominent approaches to safeguard sensitive information while enabling collaborative model training. However, DP has failed, for example, in federated learning setting~\citep{wang2021protecting}. FL assumes non-malicious participants, which is highly unlikely in real-world applications.

Linkage attacks have garnered attention in the context of privacy preservation. These attacks exploit auxiliary information to re-identify individuals within a supposedly anonymized dataset. According to~\citet{Sweeny2002}, 87\% of the US population can be identified by a limited combination of their attributes. They show how seemingly non-sensitive information can be used to re-identify individuals in a dataset. Further, they pioneered research in attribute inference attacks, where an adversary may infer values of privatized attributes from other attributes or datasets. Such attacks have been demonstrated across various domains, including healthcare~\citet{reidentmajorattack} and social networks~\citet{Gulyas2016}, underscoring the importance of robust privacy mechanisms.~\citep{Shokri2017} demonstrated that an adversary with access to a black-box model can infer whether a particular data point was used in the training set, the so-called membership inference attack. This type of attack highlights vulnerabilities in the protection of individual privacy. While membership inference attacks have raised awareness about model privacy, they assume a specific threat model and may not generalize to all scenarios. 

There has been a growing interest in balancing data privacy and the performance in different Bandits settings to better understand the fundamental trade-off imposed by privacy constraints.~\citet{Wenbo2020} studied multi-armed-bandits where users may not want to share rewards. They study this setting under local differential privacy and proposed algorithms that match the regret lower bound of their setting up to constant factors for a given local differential privacy budget. Their work shows an increase in convergence time and, thus, delayed learning induced by the additional privacy constraint.~\citet{Malekzadeh2020} developed a framework for bandit learning in a setting akin to federated learning, where multiple agents share information privately for accelerated learning. Using a clustering approach, their evaluation shows that privatized agents outperform their non-private counterparts in limited data domains while still being competitive when more data is available. These results may suggest a more complicated exploration-exploitation dilemma than in a non-private bandit setting. Privatization may aid or not significantly impact generalization, and algorithms that tailor their choice of action-taking to privatization schemes may achieve superior early-stage performance. As pointed out by~\citet{Gkountouna2022}, clustering approaches, while preserving data privacy, may negatively impact data utility, thus requiring more targeted instance-specific approaches that target sensitive attributes between records heterogeneously instead of simply clustering records with attributes that are close in value. They provide a privatization algorithm that resists machine-learning-based approaches and thus provides practical applicability in the face of contemporary attack methods. As of the current writing, there is no prior research showcasing privacy considerations exemplified in a latent bandit framework, a gap our work aims to address.

\section{Problem formulation}
In this section, we describe the recommendation system and the linkage attack scenario. This system aims to provide recommendations to users, such as music, movies, or mobile health interventions. Each user receives recommendations from an agent, a latent bandit instance in our study. The agent estimates the user's current state, such as an activity, and chooses an item (arm) from a finite set of recommendations. Users change their state in a random manner governed by a transition matrix, and the recommender system needs to adapt to these changes as quickly as possible. More formally, we define an agent as follows. % in this system in the following way.  

\textbf{The Agent: Non-stationary Latent Bandit.} % The notation we use is as follows. 
The set of arms, corresponding to items to recommend, is $\mathcal{A} = \left[K\right]$. The set of states is denoted as $\mathcal{S}$. We use capitalized letters for all random variables. The non-stationary latent bandit~\citep{Hong2020} is an online learning problem with \textit{bandit feedback}. That is, only the reward of the chosen arm is revealed to the agent. At every time step $t = 1, 2 , \dots, n$:

\begin{enumerate}%[leftmargin=+15pt]
    \item The agent chooses an arm $A_t \in \mathcal{A}$ according to its \textit{policy}, thus acting as a mapping of \textit{history} $\mathcal{H}_t = (A_1, R_1,\dots, A_{t-1}, R_{t-1})$ to arms in $\mathcal{A}$. 
    \item The environment reveals the reward $R^{A_t} \in \mathbb{R}$ according to the joint conditional reward distribution $P(R | A, S; \theta)$, parameterized by $\theta \in \Theta$, where $\Theta$ is the space of plausible reward models. Thus, only if the agent correctly identifies the state of the user, is it able to provide the right recommendation.
\end{enumerate}

The starting state $S_1$ comes from a prior distribution $P_1(s)$ and changes using a transition kernel $P(S|S_{t-1};\phi)$ with parameters $\phi$ that maps the current state to a distribution over subsequent states. The agent does not change the environment state through its choices, and only the current state determines the distribution over the next states and the rewards.

The mean reward for an arm $A_t$ in latent state $S_t$ is defined as $\mu{(A_t, S_t;\theta)} = \EX_{R\sim P(R | \cdot)}\left[R\right]$ under the model parameters $\theta$. The form of $P(\cdot | A, S; \theta)$ is not restricted and can be arbitrarily complex, but the rewards for a particular arm and state are assumed to be Gaussian distributed with mean $\mu{(A_t, S_t;\theta)}$ and variance proxy $\sigma^2$. The optimal arm in latent state $S_t$ is denoted as $A_t^* = \argmax_{A \in \mathcal{A}} \mu{(a, S_t;\theta)}$. 

\begin{figure}
     \centering
     \includegraphics[width=0.5\textwidth]{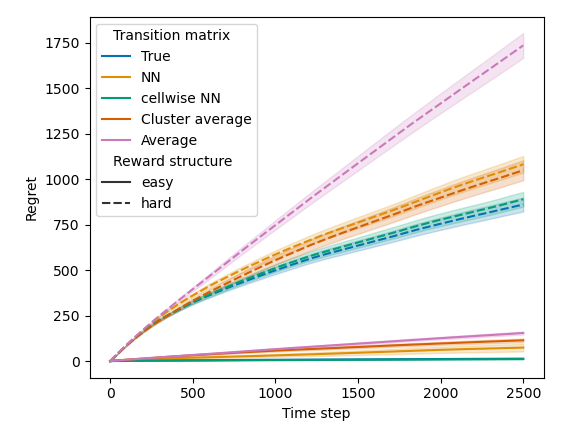}
     \caption{Hard vs. easy rewards. The accuracy of the transition matrix has diminishing returns on regret (lower is better) for easier reward structures. Regret curves are shown for different aggregation strategies on the transition matrix. For details on aggregation, see section \textit{Data anonymization}.}
     \label{fig:reward_structure}
\end{figure}

In our experiments, we focus on ``difficult'' rewards. In essence, the agent will encounter difficulty identifying the current state using only the sequence of received rewards. This will place more importance on estimating a suitable transition matrix for decision-making and show differences in reward between different strategies more clearly. 
Figure \ref{fig:reward_structure} shows an example from the Human Activity Recognition from Continuous Ambient Sensor Data (CASAS) dataset, where a hard reward structure results in significant regret differences between transition matrices.  
More specifically, what makes the rewards hard is that each state has a single optimal action with $\mu^*(\cdot) = 2$, while all the other actions have exactly the same reward, namely $\mu(\cdot) = 1.05$. For all arms, we set $\sigma = 2$. Therefore, once the user switches to the next state, an agent cannot immediately deduce the new state based on the rewards obtained by repeating the previous arm or by random exploration. Knowledge of the expected new state, based on a well-estimated individual transition matrix, is crucial for maximizing reward.

% We use this reward structure for all datasets. 

Each user has such a latent bandit agent, personalized for their needs. In our setup, we assume that the true reward distribution $P(\cdot | A, S; \theta)$ is known to the agent, while the transition kernel (transition matrix) $P(S|S_{t-1};\phi)$ is not. The agent needs to estimate the individual matrix based on trial and error. However, users collaborate to improve recommendation quality by sharing anonymized versions of these matrices.  At the same time, this opens them to a potential attack by an adversary, which we will describe next.

\begin{figure}
    \centering
   \includegraphics[width=0.7\textwidth]{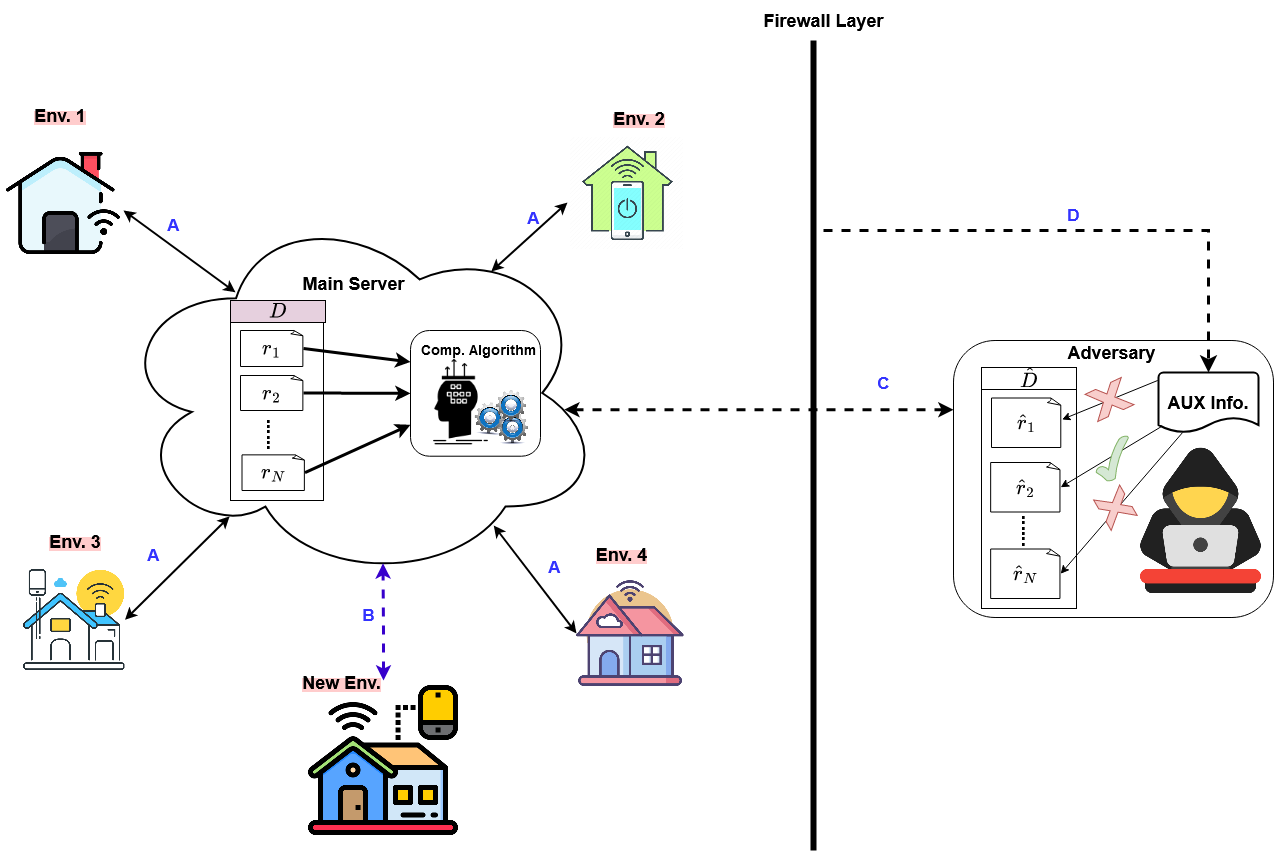}
    
\caption{A high-level representation of the recommender system and adversary model.
    %High-level schema of the recommender system and adversary model.
    }
    \label{fig:adv_model}
\end{figure}

\textbf{Sharing transition matrices and the adversary model.} The process of sharing information among users is demonstrated in Figure \ref{fig:adv_model}, describing the learning setup and the adversary. Distributing transition matrices is handled by a centralized server, which receives requests from all agents and authenticates their identities. The agents form a federation or learning group, and each member shares their anonymized transition matrix, and additional metadata, with the server. This way, a dataset $D$ is formed (\textbf{A}). Once a new agent has been added to the system, it initiates the learning process by asking the server for a suitable transition matrix. The matching is done based on the metadata, a user profile specific to each dataset. This profile is used by the server for measuring similarity across the clients and identifying the most beneficial matrix for the new arrival (\textbf{B}).
% selected by the computing algorithm on the server side based on the new agent's provided metadata (B). 

The adversary, pretending to be a new user, may communicate with the server, send fake metadata, and receive a transition matrix. An adversary may perform this several times, with different metadata, to obtain different transition matrices. Thus, it is constructing an anonymized version $\hat{D}$ of dataset $D$, containing the transition matrices of potentially all the users in the learning group (\textbf{C}). %Since the adversary possesses auxiliary information about the users in the learning group, they may attempt to match this partial data (a percentage of cell values) to full transition matrices $\hat{D}$ (D). 
%\textbf{Matching strategy of the adversary.} 
The adversary follows a similar procedure to the one described by~\citet{Narayanan2008}, attempting to match collected auxiliary information (a percentage of cell values) to full transition matrices $\hat{D}$ (\textbf{D}). First, we sample one user's transition matrix from dataset $D$. From the selected transition matrix, we provide the adversary with auxiliary information that contains the values of a fixed number of cells. We chose those cells uniformly at random. In our experiments, we vary the number of available cells to the adversary. For all runs, the adversary can access an anonymized version $\hat{D}$ of $D$.

% the data that is sent to the server to provide the agent of new users with a model to facilitate good recommendation. How these transition matrices are shared, and a potential attack vector to be exploited by an adversary are described next.  

%We describe a more detailed procedure in the experimental section.

% /////////////////////////////////////////////////////
% JOURNAL PAPER

We follow a similar procedure to the one described by~\citet{Narayanan2008}. First, we sample a transition matrix from database $D$. Then, we provide the adversary with auxiliary information that contains the values of a fixed number of randomly chosen cells from the selected transition matrix. We chose cells uniformly at random. The adversary has access to an anonymized version $\hat{D}$ of $D$. The adversary aims to identify the transition matrix in $\hat{D}$ what the auxiliary information belongs to, thus reconstructing the other cells. The process is demonstrated in figure \ref{fig:adv_model}.

\begin{figure}
    \centering
   \includegraphics[width=0.8\textwidth]{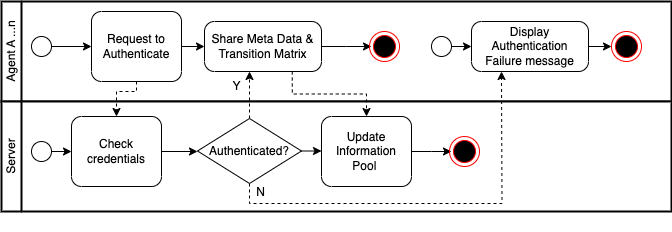}
    \caption{The process diagram for forming alliances enables rapid learning and knowledge sharing among various agents through a centralized server.}
    % {The alliance process diagram for fast learning and sharing knowledge between different agents in a centralized scheme.}
    \label{fig:process1}
\end{figure}
The process diagram shown in Figure \ref{fig:process1} illustrates the steps involved in allying various agents possessing varying levels of expertise to promote fast learning and knowledge sharing within the group and for a new agent. The process is handled by a centralized server, which receives requests from all agents and authenticates their identities. Once authenticated, the agents form a federation or learning group, and each member must share their transition matrix and environment metadata with the server.
\begin{figure}
    \centering
   \includegraphics[width=0.8\textwidth]{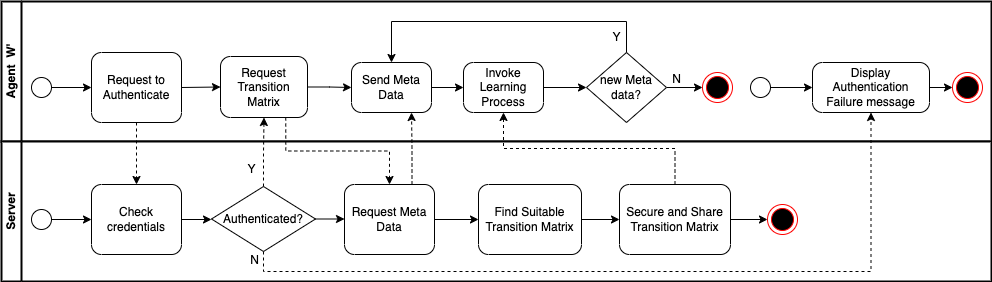}
    \caption{The process diagram of new agents joining the alliances.}
    \label{fig:process2}
\end{figure}
 As a result, the server must authenticate any new agents before they join the group and share their environment metadata to enable them to leverage the knowledge of other members and avoid starting from scratch, as shown in Figure \ref{fig:process2}. Once the agent has been added to the system, it can initiate the learning process by obtaining a suitable transition matrix selected by the computing algorithm in the server side based on the new agent's provided metadata. Additionally, if any change occurs in any agent's environment, such as adding a new sensor or feature. In that case, the agent can update their metadata and request another transition matrix that is more appropriate for its updated settings.

\begin{figure}
    \centering
   \includegraphics[width=0.8\textwidth]{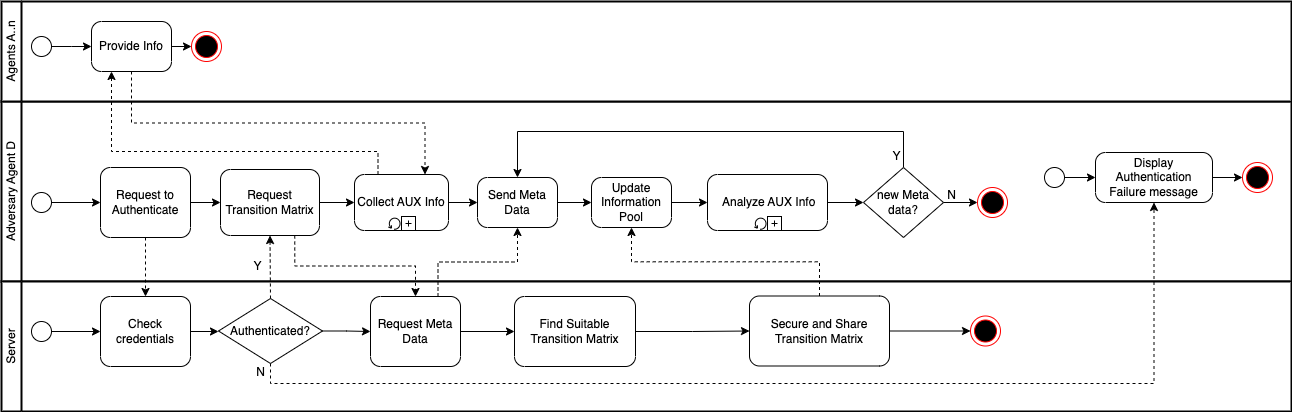}
    \caption{Flow diagram of the adversary.}
    \label{fig:process3}
\end{figure}

 In certain scenarios, an adversarial entity assumes the role of a new participant seeking to infiltrate the federation with intentions such as uncovering the original data flow among participants, disrupting the entire system, or exploiting it for alternative purposes. Figure \ref{fig:process3} outlines the sequence of actions this adversary uses to deceive the system. The adversary endeavors to integrate into the federation under the guise of a regular participant seeking knowledge sharing instead of starting anew in a separate environment.

The process begins with the adversary sending authentication requests, aiming to gain the server's trust. Once this trust is established, the adversary is admitted into the system. Subsequently, the adversary initiates requests for a transition matrix, providing the requisite metadata. Simultaneously, this fraudulent participant allocates time to accumulate publicly available auxiliary information within the system concerning other participants. This acquired information can potentially be leveraged to extract sensitive data from the transition matrix about other participants. The adversary maintains a repository for this information, which can be continually updated.

To maximize the acquisition of the transition matrix, the adversary alters their identity and environmental settings, repeating the steps mentioned earlier until their objectives are met.

%///////////////////////////////////////////////////

\section{Data anonymization}\label{sec:data_anym}

\textbf{Additive noise and lossy compression.} In this paper, we investigate two popular methods for directly changing the values within individual transition matrices. The first method adds noise to sensitive information via a Laplace mechanism. For noise addition, we formally define the noisy version of the transition matrix as
\begin{equation}
    \hat{\mathbf{r}} = \mathbf{r} + \mathbf{w}\cdot Z(\epsilon), \forall \mathbf{r} \in D,
\end{equation}
where $Z$ is a random draw from a Laplace distribution with mean $0$ and $\epsilon$ variance, and  $\mathbf{w}$ represents the weight of the added noise to individual cells and can be chosen arbitrarily.

%For a formal definition of $\mathbf{w}$ see equation \ref{eq:entropy_weight} later where we define ADS Identifiability.
The second method is truncated singular value decomposition (tSVD), widely used for matrix approximation. By considering only the top-$k$ eigenvalues, tSVD compresses the information in the transition matrix while preserving essential information for approximate reconstruction. Specifically, tSVD solves the problem of approximating the transition matrix $\mathbf{r}$ with another matrix $\Tilde{\mathbf{r}}$ by minimizing the Frobenius norm $||\mathbf{r}-\Tilde{\mathbf{r}}||_F$ subject to the constraint that $\operatorname{rank}(\Tilde{\mathbf{r}})$ is equal to $k$~\cite{svd}. Importantly, tSVD also offers the benefit of removing identifying information. Cells, including those corresponding to unique user characteristics, are not reconstructed accurately if they do not affect the Frobenius norm significantly during optimization, increasing the difficulty of identifying users based on these cells.

\textbf{Aggregation strategies.} Besides adding noise to individual records, we investigate different methods of aggregating transition matrices across users before providing them to the agent for decision-making. We specifically focus on averaging methods, that is, calculating elementwise means of the transition matrices of several users. Averaging has the effect of hiding user-specific information, thus making linkage attacks harder. We explore three methods: nearest neighbor (NN), cluster average (Cluster average), and global average (Average). Further, we investigate some variants for the experiments. Specifically, we evaluate the impact (in terms of regret performance and privacy) of choosing a ``more distant'' transition matrix using the second nearest neighbor (Second NN). For clustering, we analyze the performance gain against the weakening of anonymization by doubling (2x Cluster average) and tripling (3x Cluster average) the number of clusters; more clusters mean fewer users per cluster, and, thus, more accurate transition matrix reconstructions. 

We more formally define the cluster strategy as follows. Given records $r = \{\mathbf{r_1}, \mathbf{r_2}, \dots, \mathbf{r_N}\} \in \mathbb{R}_{\geq 0}^{d \times d}$ and $c$ clusters $\{c_1, c_2, \dots, c_C\} \in \mathbb{Z}^+$, we assign each record $i$ to one cluster $\pi(i) \in \{1, 2, \dots, C\}$. We denote $|c_{\pi(i)}|$ as the cardinality (number of users) of cluster $c_{\pi(i)}$ of record (user) $i$. With slight abuse of notation, we abbreviate $|c_{\pi(i)}|$ as $|c_{i}|$.

The similarity between users for the nearest neighbor and clustering computations is not based on the values of the cells in the transition matrices but rather on user profiles (metadata) since, in real-world applications, the transition matrix of new users is unknown, at least initially. We compute a user profile for each dataset that approximately encapsulates the user's behavior. We describe the details in the \textit{Experimental setup} section.

%\textbf{Matching strategy of the adversary.} We follow a similar procedure to the one described by \citet{Narayanan2008}. First, we sample a transition matrix from database $D$ corresponding to a random user. From the selected transition matrix, we provide the adversary with auxiliary information that contains the values of a fixed number of chosen cells. We chose cells uniformly at random. For our experiments, we vary the number of available cells to the adversary. For all runs, the adversary has access to an anonymized version $\hat{D}$ of $D$. The adversary aims to identify the transition matrix in $\hat{D}$ that corresponds to the provided auxiliary information. 

\section{Metrics and algorithms}

\textbf{Regret.} For the performance metric, we use regret, the most common metric in bandit learning, defined as follows. For a fixed latent state sequence $s_{1:n} \in \mathcal{S}^n$, the expected n-step regret is defined as
\begin{equation}
    \mathcal{R}(n, \phi, S_{1:n}) = \EX\Bigg[\sum_{t=1}^n\mu(A_t^*, S_t;\theta) - \mu(A_t, S_t;\theta)\Bigg],
\end{equation}
where the expectation is taken over the agent's randomness in arm choice. We consider the Bayes regret, computing the n-step regret as an expectation over latent state randomness, defined as $\mathcal{BR}(n;\theta;\phi) = \EX_{S_{1:n} \sim \phi} [\mathcal{R}(n, \phi, S_{1:n}) | \theta, \phi]$.

\textbf{Probability of de-anonymization.} In the following, we define whether a data record is successfully de-anonymized given auxiliary information depending on the performance of the scoreboard algorithm~\cite{Narayanan2008}. Scoreboard is a generic algorithm based on nearest-neighbor matching. The principal procedure, with some modifications, is as follows:

\begin{enumerate}% [leftmargin=+15pt]
    \item Compute the score $\operatorname{Score}(aux, \mathbf{r}')$ according to the average similarity of  attributes between the candidate record $\mathbf{r}' \in \hat{D}$ and the auxiliary information as $$\operatorname{Score}(aux, \mathbf{r}')=\frac{1}{|\sup(aux)|}\sum_{i\in \sup(aux)} \operatorname{Sim}(aux_i, \mathbf{r}'_i).$$
    The support on auxiliary information $aux$ is denoted as $\sup(aux)$ and corresponds to the cells available to the adversary for comparison. Only the cells in $aux$ are used to compute the similarity to records in $\hat{D}$.  
    \item Compute the matching set $D'$ as $D' = \{\mathbf{r}' \in \hat{D} : \operatorname{Score}(aux, \mathbf{r}') = \max_{\mathbf{r}' \in \hat{D}} {\operatorname{Score}(aux, \mathbf{r}')}\}$.
    \item Sample uniformly a candidate record from $D'$.
\end{enumerate}
In this paper, $\operatorname{Sim}(\cdot)$ computes $L_1$ distance, with some fixed constant $\alpha$, between cells in $aux$ and the corresponding cells in a candidate record in $\hat{D}$ as $\operatorname{Sim}(aux_i, \mathbf{r}'_i) = (aux_i-\mathbf{r}'_i) < \alpha$. 

The probability of de-anonymization for a dataset $D$ is computed as $$\frac{1}{N} \sum_N \mathbbm{1}(\mathbf{r}_i = \mathbf{r}'_i),$$
% The probability of de-anonymization for a dataset $D$ is computed as 
% \begin{equation}
%     \frac{1}{N} \sum_N \mathbbm{1}(\mathbf{r} = \mathbf{r}').
% \end{equation}
where $N$ is the number of records (one record per user) in the dataset. The probability of de-anonymization is, in other words, the average number of records $r'$ correctly matched with the corresponding record $r$ using $aux$.

To interpret our experimental results correctly, it is important to note that, for clustering strategies, we only share the average transition matrix of a cluster, not the transition matrices of individual users. Thus, $\hat{D}$ constitutes a set of cluster averages. Users of the same cluster have the same cluster average in $\hat{D}$, plus some possible noise individual to each user. As a result, the adversary can only hope to match the auxiliary information to a cluster as a whole. To calculate the probability of de-anonymization, we consider the success rate of matching the auxiliary information to a particular cluster, as well as the size of the cluster. For instance, if a cluster has five users, the probability of de-anonymization is $0.2$ for a correct match and $0$ otherwise. More formally, the level of de-anonymization using an aggregate transition matrix is denoted (interpreting $\mathbf{r} _i$ and $\mathbf{r} _i'$ now as cluster averages instead of the transition matrix of a particular user) as $$\frac{1}{N} \sum_{i=1}^N \mathbbm{1}(\mathbf{r}_i = \mathbf{r}_i')\cdot \frac{1}{|c_i|}.$$ 
% Note that record $\mathbf{r} _i$ and candidate record $\mathbf{r} _i'$ are now interpreted as cluster averages instead of the transition matrix of a particular user.

\textbf{ADS privacy} is a popular metric for estimating the level of privacy provided in a dataset. It is a metric used to minimize a corresponding loss in ADS-GAN, maximizing privacy intelligently while retaining performance in downstream tasks using the anonymized dataset~\cite{Yoon2020}. In this study, we investigate if the metric accurately captures the probability of de-anonymization via linkage attacks. We compute the ADS metric for different Laplace noise levels and the number of cells of auxiliary information available to the adversary across three datasets. ADS privacy is defined as
\begin{equation}\label{eq:adsid}
\mathcal{I}(\mathcal{D}, \mathcal{\hat{D}}) = \frac{1}{N}\sum_{i=1}^N[\mathbbm{1}(\hat{k}_i < k_i)] < \epsilon,
\end{equation}
where $\mathbbm{1}$ is the indicator function% and $k_i$ and $\hat{k}_i$ are defined as
, $k_i = \min_{\mathbf{r}_i \in \mathcal{D} \setminus \mathbf{r_j}} || \mathbf{w} \cdot (\mathbf{r}_i-\mathbf{r}_j)||$, and $\hat{k}_i = \min_{\mathbf{\hat{r}}_j \in \mathcal{\mathcal{\hat{D}}}} || \mathbf{w} \cdot (\mathbf{r}_i-\mathbf{\hat{r}}_j)||$.
%\begin{align*}
%\centering
%k_i &= \min_{\mathbf{r}_i \in \mathcal{D} \setminus \mathbf{r_j}} || \mathbf{w} \cdot (\mathbf{r}_i-\mathbf{r}_j)||, \\
%\hat{k}_i &= \min_{\mathbf{\hat{r}}_j \in \mathcal{\mathcal{\hat{D}}}} || %\mathbf{w} \cdot (\mathbf{r}_i-\mathbf{r}_j)||.
%\end{align*}

One typically uses the measure \ref{eq:adsid} to assess whether synthetic records (created by adding noise to the original transition matrices) differ sufficiently from the original records. It is based on the concept of $\epsilon$-identifiability, which means that the probability of a synthetic observation being closer to the original observation than another original observation is less than $\epsilon$. The value of $\epsilon$ determines the proportion of users that synthetic observations cannot identify because they differ enough from the real observations. Specifically, if the proportion of unidentifiable users is $\geq 1 - \epsilon$, the synthetic observations provide sufficient privacy protection. When $\epsilon$ equals 0, the data is perfectly non-identifiable~\citep{Yoon2020}.
%The measure determines if synthetic records (noised transition matrices in this paper) are ``different enough'' from original records to protect privacy. In the context of the measure, $\epsilon$-identifiability means that the probability of a synthetic observation being closer to the original observation than another original observation is less than $\epsilon$. In other words, $\geq$ 1 - $\epsilon$ is the proportion of users  not identifiable by synthetic observations because they are ``different enough'' from the real observations. $0$-identifiability means that the data is perfectly non-identifiable~\cite{Yoon2020}. 

%The weight vector $\mathbf{w}$ in this paper is defined as the inverse of the discrete entropy for each cell in the transition matrix. More formally, the discrete entropy is defined as $H(X_i) = \sum_{x_i \in \mathcal{X}_i}P(X_i = x_i) \log P(X_i = x_i)$.

%Where $X_i$ is a random variable for a cell in the transition matrix, and the weight vector $\mathbf{w}$ defined as the inverse of the discrete entropy of each cell as described earlier is section \ref{sec:data_anym}.

\begin{algorithm}[!b]
\caption{mTS~\citep{Hong2020}}
\label{alg:ps}
\begin{algorithmic}[1]
\State \textbf{Input:} 
\State \;\;\;\; Model parameters $\theta, \phi$, 
\State \;\;\;\; Prior over initial latent state $P_1(s)$
\For{$t=1,2,\dots,n$}
    \State Sample $B_t \sim P_t$
    \State Select the best arm according to $B_t$: $A_t = \argmax_{a \in \mathcal{A}}  \mu(A, B_t; \theta)$ and observe $R_t$ 
    \State Update belief state posterior $P_{t+1}(S_{t+1}) \propto \sum_{S_t \in \mathcal{S}} P_t(S_t)P(S_{t+1} | S_t;\phi) P(R_t | A_t, S_t;\theta)$
\EndFor
\end{algorithmic}
\end{algorithm}

\textbf{Model-based Thompson Sampling (mTS)}.
We selected Model-based Thompson Sampling (mTS) as the recommendation algorithm, a posterior sampling technique demonstrating state-of-the-art performance in the latent bandit setting according to previous research~\citep{Galozy2023, Hong2020}. mTS uses a posterior update rule and selects an arm based on its probability of being optimal given the history $\mathcal{H}_t$, which is expressed as $\mathbb{P}(A_t = a | \mathcal{H}_t) = \mathbb{P}(A^*_t = a | \mathcal{H}_t)$. To select an arm, mTS samples a state $B_t \in \mathcal{S}$ from its posterior distribution over the belief state $P_t(s)$ and chooses the arm $A_t$ with the highest mean reward given the sampled $B_t$, as $A_t = \argmax_{a \in \mathcal{A}} \mu(a, B_t;\theta)$. After the reward $R_t$ is received for the chosen arm, the posterior of the belief state $P_{t+1}$ is formed using the Bayes rule. The algorithm's pseudo-code is shown in Algorithm \ref{alg:ps}. In our experiments, mTS has access to the reward predictor $P(R_t|A_t, S_t; \theta)$, a reasonable assumption given that we may use prior collected data to construct the predictor. The available data allow mTS to quickly provide good recommendations without learning from scratch~\citep{Hong2020}.

\section{Evaluation}

\subsection{Experimental setup}

%SUP MATERIAL
%For easy rewards, we again set $\mu^* = 2$, but the sub-optimal arms have a unique mean reward value spanning a linear space from $2$ to $0$ with a step size that divides this space into $|\mathcal{S}|$ equally sized chucks. Furthermore, we set $\sigma = 0.01.$   

\textbf{Datasets and transition matrices.}
We run experimentation on three datasets, each containing user activity data. From the activity data, we extract states and state transition probabilities. For all datasets, the time in seconds in each state defines the diagonal elements of the transition matrix. In contrast, the off-diagonal elements are computed from the frequency count of observed transitions between states. In the following, we describe what constitutes the states in each dataset.  

\begin{itemize}%[leftmargin=+10pt]
    \item \textbf{Human Activity Recognition from Continuous Ambient Sensor Data (CASAS)} dataset was collected from several apartments that contain ambient sensors~\citep{casa_dataset}. 41 activities in the dataset constitute the states in the transition matrix. The transition matrix is of size $41\times41$. 
    \item \textbf{Endomondo} dataset was collected from the Endomondo app~\citep{fitRec}. The types of sports activities constitute the states in the transition matrix. The transition matrix is of size $39\times39$. 
    \item \textbf{Fitbit} dataset was collected through a distributed survey conducted by Amazon Mechanical Turk based on participants' responses~\citep{fitbit}. The four levels of activity intensity constitute the states. The transition matrix is of size $4\times4$.  

\end{itemize}

\begin{figure}
    \centering
    \includegraphics[width=0.7\textwidth]{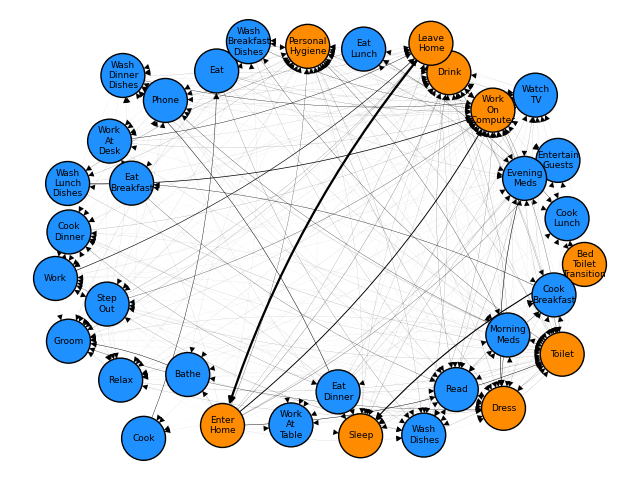}
    \caption{Example transition graph of a user in the CASAS dataset. The ten most common states in the dataset, among all users, are marked in orange.}
    \label{fig:trans_graph_casa}
\end{figure}

Different anonymization strategies, i.e., additive noise, averaging, or tSVD, can add varying amounts of total noise to the transition matrix. To ensure a fair comparison in our experiments, we $L_2$-normalize the transformed matrix. Moreover, tSVD or noise addition may produce negative values in some cells, which we truncate to zero. %This normalization allows us to standardize the scale of the transformed matrix across different transformation methods, which facilitates a more accurate comparison of their performance. 
Subsequently, the transition matrix is row-wise normalized so it sums to one.  Missing states (rows and columns of the transition matrix with only zero values) are imputed with a low non-zero value. Specifically, the environment is assumed to have a uniform probability of transitioning to any other state from the missing one. An example of a transition graph for a user in the CASAS dataset can be seen in Figure \ref{fig:trans_graph_casa}.

\textbf{User profiles.} For clustering and nearest neighbor computation, the user similarity is estimated using precomputed user profiles for each dataset, generated in the following way:

\begin{itemize}%[leftmargin=+10pt]
    \item \textbf{CASAS.} An adjacency matrix of all apartment sensors is created from the sensor activations of each user. Transitions between activations of sensors are recorded, and the corresponding cell in the adjacency matrix is filled with 1, indirectly creating a layout graph of the apartment. The graphs are embedded into a 128-dimensional vector space via Graph2Vec~\citep{narayanan2017} using a parallel implementation in the karateclub python package~\citep{karateclub} with standard parameters.
    \item \textbf{Endomondo.} We extract the average time spent in an activity, heart rate, and distance computed over all activities. Additionally, the frequency for each activity is extracted. This results in a  43-dimensional vector constituting the user profile.
    \item \textbf{Fitbit.} User profiles are created from hourly activity intensities resulting in a 24-dimensional vector containing the average intensities for each hour of the day over all days for a user.
\end{itemize}

We first cluster the user profiles in each dataset using the k-means algorithm from the scikit-learn python package~\citep{scikit-learn}. We explored various numbers of clusters for each dataset using unsupervised clustering metrics\footnote{Scores used in this study include Akaike information criterion, Bayesian information criterion, Davies Bouldin score, and Calinski Harabasz score.} without a clear optimum in the number of clusters. Thus, we have chosen a low but reasonable number of clusters: eight for Endomondo and Fitbit, and seven for CASAS. 
%for journal

\begin{figure}
     \centering
     \begin{subfigure}[b]{0.32\textwidth}
         \centering
         \includegraphics[width=\textwidth]{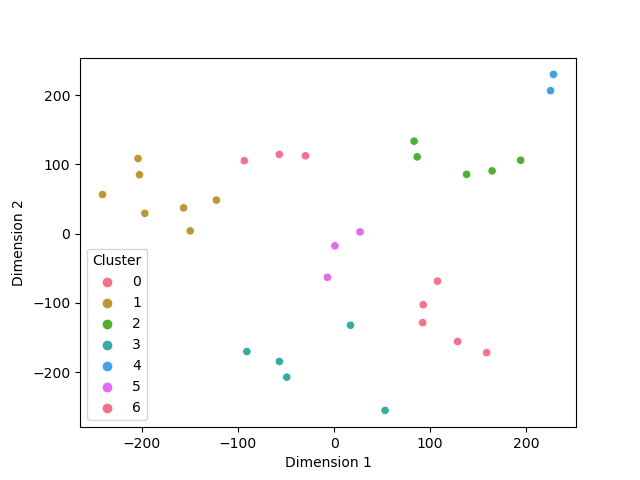}
         \caption{CASAS}
 
     \end{subfigure}
     \hfill
     \begin{subfigure}[b]{0.32\textwidth}
         \centering
         \includegraphics[width=\textwidth]{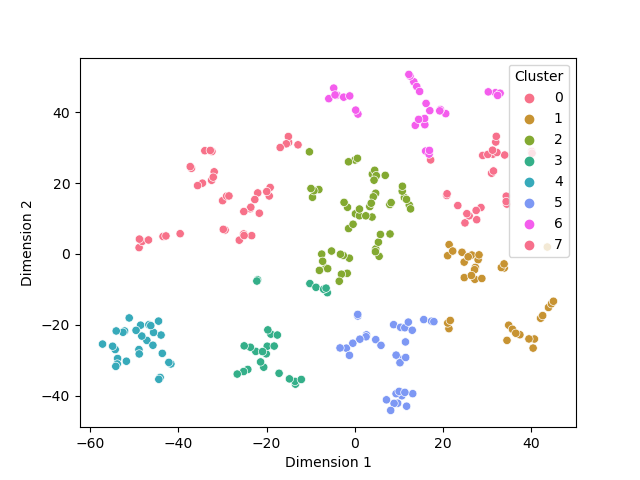}
         \caption{Endomondo}
        
     \end{subfigure}
     \hfill
     \begin{subfigure}[b]{0.32\textwidth}
         \centering
         \includegraphics[width=\textwidth]{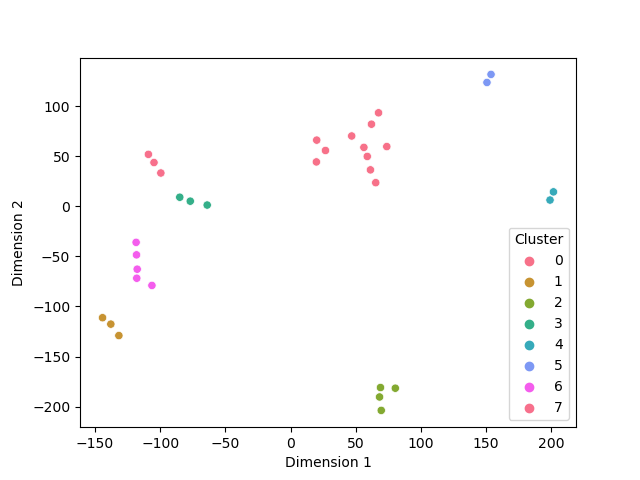}
         \caption{Fitbit}
     \end{subfigure}
        \caption{Clustering of the 2D-TSNE embedding.}
        \label{fig:clusters}
\end{figure}

Due to the high dimensionality of the user profile data, we first transform the user profiles using t-SNE~\citep{t-sne} to embed them into a separate 2D space before clustering. The resulting clusters are shown in figure \ref{fig:clusters}. 
It is worth noting that t-SNE, as an embedding technique, is not guaranteed to preserve intercluster distances, which may hurt the performance of mTS. Nevertheless, we have not observed this as being an issue in practice.
% not shown any appreciable impact on our experimental results, and the conclusions still hold. 

%Journal
%Similarly, it is essential to consider the balance between perfect and imperfect clustering regarding privacy. Perfect clustering may reveal too much information about individuals or groups, which can compromise privacy. Conversely, imperfect clustering, which introduces noise and ambiguity, can make distinguishing between clusters more challenging for the adversary. 

\subsection{Run parameters and computational resources} For the ADS-metric study, we chose the number of auxiliary cells for the datasets CASAS and Endomondo between $1$ to $100$ in $10$ cell increments and between $1$ and $15$ in one-cell increments for Fitbit. In the privacy-performance trade-off experiments, we choose the number of cells available showing the most significant change in de-anonymization probability for a given noise level: $91$ cells (CASAS and Endomondo) and $14$ cells (Fitbit), respectively. We compute the regret for the time horizon of $2500$ timesteps as an average of over $500$ runs. For both experiments, we vary the variance of the Laplace noise in the interval $[0, 3)$ (CASAS and Endomondo) and $[0, 3\times10^-4)$ (Fitbit) divided into $50$ evenly spaced numbers. For tSVD the number of $top-k$ eigenvalues varies from $41$ (CASAS), $39$ (Endomondo) and $4$ (Fitbit) to $1$. The probability of de-anonymization is computed as an average of $100$ runs, with a random user chosen from $D$ in each run. We do this to account for the randomness of added noise. For the scoreboard algorithm, we choose $\alpha = 0.001$ for all experiments. Further, for the additive noise and the scoreboard algorithm, we define 

\[\mathbf{w} = (w_1, \dots, w_d) = \big(1/{H(X_i)}, \dots, 1/{H(X_d)}\big),\] with $w_i \in [0, \infty)$ and entropy \[H(X_i) = \sum_{x_i \in \mathcal{X}_i}P(X_i = x_i) \log P(X_i = x_i).\] The inverse entropy encodes the concept of ``uniqueness'' regarding the cell's value distribution in the dataset. The more unique the cell value of a user's transition matrix, the more noise is added. 

For a specific parameter setting, regret performance experiments over $500$ runs take about 20 minutes on a Ryzen 9 5950x CPU with 32GB RAM, single-threaded. The computation of de-anonymization probability takes a fraction of a second on the same system.    

%Journal
%Suppose most users have a cell in the auxiliary information with similar values. In that case, it is not a good choice for identifying individuals, and much noise may not be necessary to privatize the cell effectively.

\subsection{Results}

\begin{figure}
     \centering
     \begin{subfigure}[b]{0.45\textwidth}
         \centering
         \includegraphics[width=\textwidth]{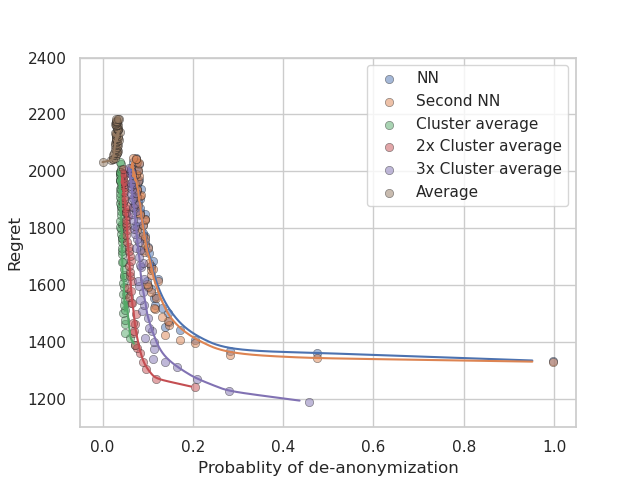}
         \caption{Laplace noise}
         \label{fig:CASAS_laplace}
     \end{subfigure}
     \hfill
     \begin{subfigure}[b]{0.45\textwidth}
         \centering
         \includegraphics[width=\textwidth]{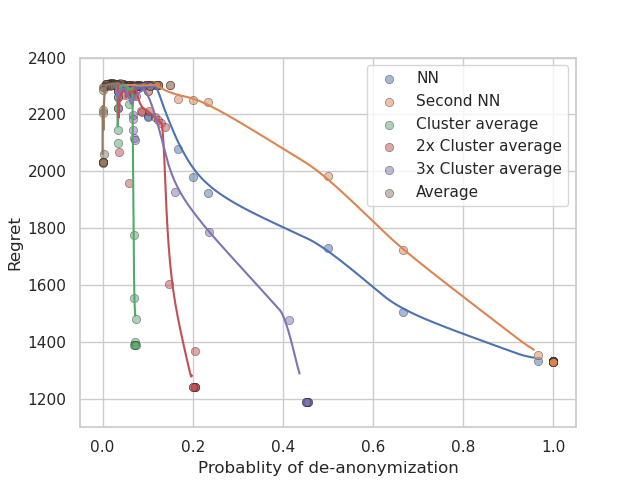}
         \caption{tSVD transformation}
         \label{fig:CASAS_svd}
     \end{subfigure}
     \caption{De-anonymization vs. regret for the CASAS dataset.}
     \label{fig:CASAS_regret_privacy}
\end{figure}

The impact of Laplace noise and tSVD on privacy and regret was investigated in conjunction with aggregation and nearest-neighbor methods. Figure \ref{fig:CASAS_regret_privacy} shows the results for the CASAS dataset, demonstrating that clustering offers a favorable balance between privacy and regret, compared to nearest-neighbor approaches at a given privacy level. Clustering significantly enhances privacy, even if no noise is added to the cluster average; and in combination with noise, it outperforms all the other methods. In particular, employing fewer clusters (e.g., seven) yields comparable performance to the nearest-neighbor approach without any noise, while achieving considerably higher (almost optimal) privacy. More clusters, on the other hand, % improves the performance of mTS if higher privacy is not required. 
offers a further significant benefit in terms of lower regret, at modest privacy sacrifices.
% greater benefit in terms of privacy, per 
The suboptimal performance of the nearest neighbor transition matrix may result from overfitting, whereas averaging over multiple users mitigates this issue without compromising privacy substantially.

Laplace noise clearly outperforms tSVD in terms of privacy and regret control across all datasets. tSVD appears to alter the transition matrix in a particularly harmful way, leading to significant regret impact even at similar privacy levels to Laplace noise. This effect is evident in the comparison of NN and Second NN regret in figure \ref{fig:CASAS_svd}. Unlike Laplace noise, tSVD amplifies the differences between NN and Second NN, significantly affecting the regret.

Surprisingly, this relationship between the number of clusters and the probability of de-anonymization does not hold for the Endomondo dataset, where increasing the number of clusters does not significantly impact privacy while significantly reducing regret. Apparently, perfect precision in the transition matrices is not necessary for high recommendation quality. We notice that higher numbers of clusters allow for an increase in privacy level compared to the baseline of eight clusters, providing the best trade-off between privacy and performance. 

%Adding Laplace noise to the aggregation process allows for finer tuning of privacy and regret than tSVD. Using tSVD alters the transition matrix enough to significantly impact regret for similar privacy levels compared to Laplace noise. The impact on performance can more clearly be seen if we compare the regret of the NN vs. the Second NN in figure \ref{fig:CASAS_svd}. tSVD exacerbates the differences between the NN and second NN enough to impact performances significantly at the same privacy level, a behavior not observed for the Laplace noise case, where the Second NN may provide a slight but measurable performance advantage. 

\begin{figure}
     \centering
     \begin{subfigure}[b]{0.48\textwidth}
         \centering
         \includegraphics[width=\textwidth]{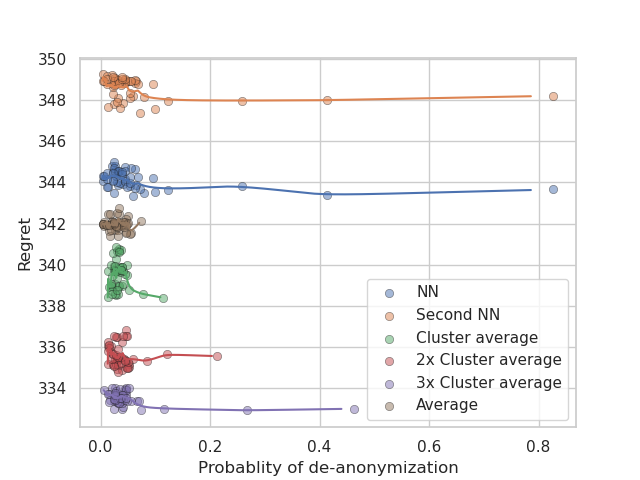}
         \caption{Laplace noise}
         
     \end{subfigure}
     \hfill
     \begin{subfigure}[b]{0.48\textwidth}
         \centering
         \includegraphics[width=\textwidth]{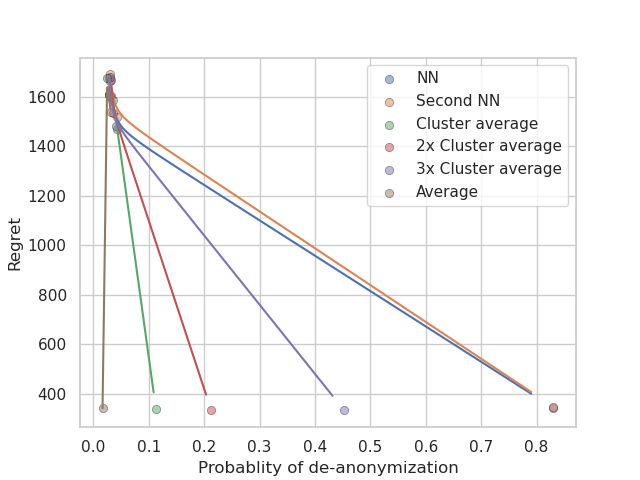}
         \caption{tSVD transformation}
         
     \end{subfigure}
     \caption{De-anonymization vs. regret for the Fitbit dataset.}
     \label{fig:fitbit_regret_privacy}
\end{figure}

Somewhat unintuitively, however, the general privacy-regret trade-off may not apply to every dataset. For example, in the case of the Fitbit dataset, privacy and performance do not have to be balanced significantly at all. As illustrated in figure \ref{fig:fitbit_regret_privacy}, high levels of privacy can be achieved without sacrificing performance using Laplace noise and a cluster strategy.

These results highlight the main takeaway of our study. For a given dataset, a combination of aggregation strategies and noise allows for more flexibility that is not achievable by either aggregation or noise alone. For example, in the range between $0.1$ and $0.3$ of de-anonymization probability in the CASAS dataset (figure \ref{fig:CASAS_laplace}), one may opt for cluster averages using double the number of clusters ($14$ versus $7$) to achieve the best performances. One might increase this number even further to achieve even better performance but at the consequences of sacrificing privacy. In general, no strategy we have investigated achieves the best privacy-performance trade-off over the complete range of de-anonymization probabilities. Thus, a combination of noise and aggregation proves necessary to assess for a given privacy level.

\begin{figure}
     \centering
     \begin{subfigure}[b]{0.48\textwidth}
         \centering
         \includegraphics[width=\textwidth]{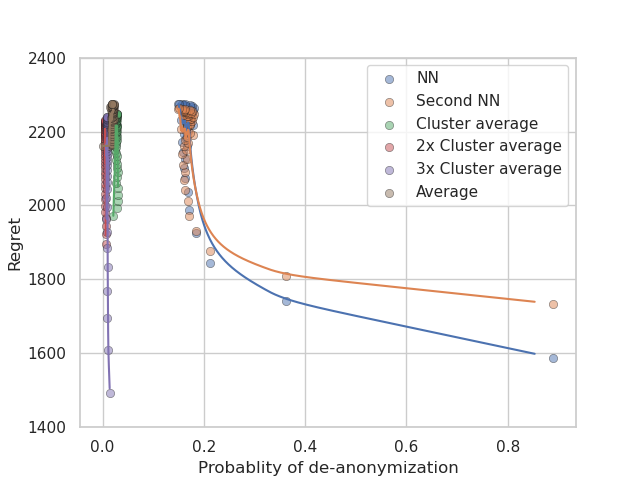}
         \caption{Laplace noise}
         
     \end{subfigure}
     \hfill
     \begin{subfigure}[b]{0.48\textwidth}
         \centering
         \includegraphics[width=\textwidth]{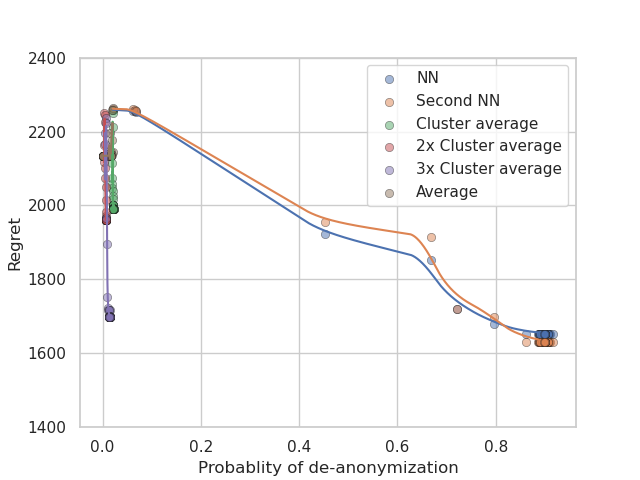}
         \caption{tSVD transformation}
     \end{subfigure}
     \caption{De-anonymization vs. regret for the Endomondo dataset.}
     \label{fig:endomundo_regret_privacy}
     
\end{figure}

\begin{figure}
     \centering
     \begin{subfigure}[b]{0.32\textwidth}
         \centering
         \includegraphics[width=\textwidth]{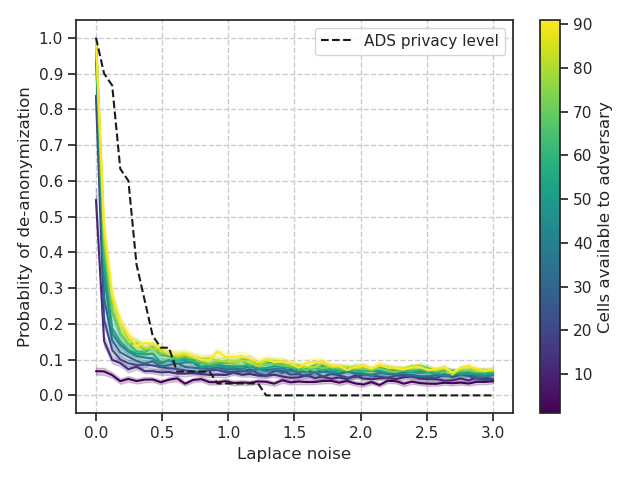}
         \caption{CASAS. Chance de-anonymization: 3.3\%}
 
     \end{subfigure}
     \hfill
     \begin{subfigure}[b]{0.32\textwidth}
         \centering
         \includegraphics[width=\textwidth]{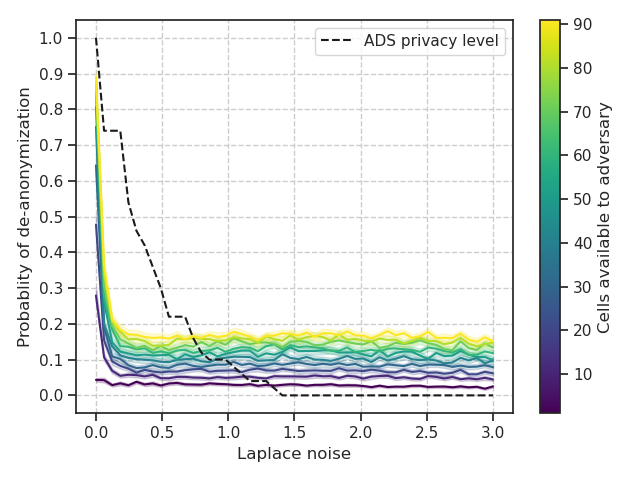}
         \caption{Endomondo. Chance de-anonymization: 2.0\%}
        
     \end{subfigure}
     \hfill
     \begin{subfigure}[b]{0.32\textwidth}
         \centering
         \includegraphics[width=\textwidth]{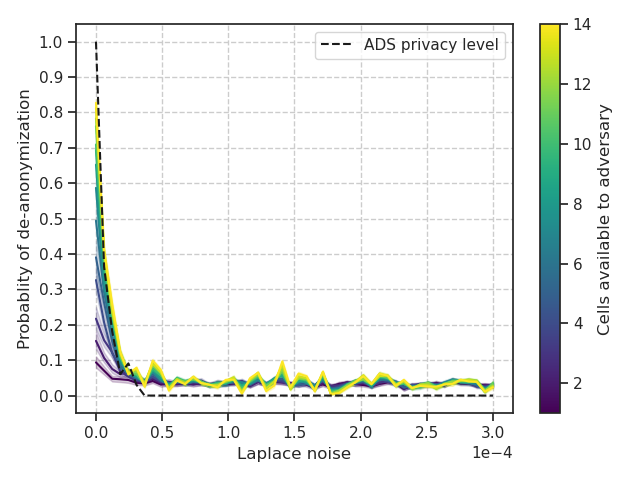}
         \caption{Fitbit. Chance de-anonymization: 3.0\%}
     \end{subfigure}
        \caption{Probability of de-anonymization and ADS privacy level vs Laplace noise.}
        \label{fig:ADS_privacy}
\end{figure}

\textbf{ADS privacy level versus de-anonymization.} The probability of de-anonymization versus level of Laplace noise is shown in figure \ref{fig:ADS_privacy}. Interestingly, we found that the privacy levels estimated by the ADS metric did not correspond well to the probability of de-anonymization. This has significant implications for privacy in cases where synthetic datasets are generated using the ADS metric for optimization. Specifically, the privacy levels may be overestimated for higher noise levels.
Consequently, the ADS metric tends to zero, leading to a false sense of security for the data owners. On the other hand, via ADS, we may also underestimate the level of privacy for low noise levels. The ADS metric tends to one resulting in the addition of more noise than required for a given de-anonymization level, potentially negatively affecting performance. It might be reasonable that ADS-GAN learns a transformation that mimics a combination of additive noise and aggregation. Still, given the definition of the metric, there is no guarantee. Further, since the anonymized dataset is not analyzed in terms of performance during training, a feedback mechanism is missing that would encourage such tailored noise. These findings highlight the need to consider the attack type instead of a ``one-size-fits-all'' approach in loss-function design.

\section{Limitations}
We did not investigate contemporary end-to-end learning schemes, such as generative models like ADS-GAN~\citep{Yoon2020} that promise to be effective data anonymization techniques. These methods might alleviate some of the concerns raised in this paper. Still, as mentioned above, these end-to-end learners often rely on objective functions that are not well aligned with common attack patterns, and thus have yet to prove effective in real-world scenarios.

Our analysis is limited to a fairly simple scoreboard algorithm for deanonymization. There may exist better algorithms with stronger deanonymization capabilities. Our research illustrated that even a simple method can lead to a significant overestimation of the level of privacy, which would only increase with more sophisticated algorithms.

Furthermore, we did not examine machine learning contexts other than bandits, such as supervised or unsupervised learning. Nevertheless, results and findings from this research are expected to transfer directly to these settings as well, since they ultimately concern data privacy irrespective of how the data is used or what specific models are trained. It is not clear to what extent privacy would be compromised in specific settings, but the general trend we observe holds. 

\section{Conclusions}

In this paper, we conducted experiments to evaluate the effectiveness of different privitization strategies, using a latent bandit setting. Our findings show that random noise added to individual records may not be a satisfactory solution for the privacy-performance trade-off. Instead, aggregations generally provide a better balance between ensuring privacy and maintaining performance for downstream tasks. There seems to be no single technique that solves the trade-off the best for a given desired level of privacy. Further, we found that one-size-fits-all metrics optimized for privacy, like the ADS-GAN metric, do not correspond well to the probability of de-anonymization in linkage attacks. Our paper highlights the need to evaluate privatization techniques on actual attack scenarios commonly employed in the real world, especially considering the abundance of useful auxiliary information collected in the wild that may render such techniques less effective.

\bibliographystyle{unsrtnat}
\bibliography{refs}  

\iffalse
\section*{References}

References follow the acknowledgments. Use unnumbered first-level heading for
the references. Any choice of citation style is acceptable as long as you are
consistent. It is permissible to reduce the font size to \verb+small+ (9 point)
when listing the references. {\bf Remember that you can use more than eight
  pages as long as the additional pages contain \emph{only} cited references.}
\medskip

\small

[1] Alexander, J.A.\ \& Mozer, M.C.\ (1995) Template-based algorithms for
connectionist rule extraction. In G.\ Tesauro, D.S.\ Touretzky and T.K.\ Leen
(eds.), {\it Advances in Neural Information Processing Systems 7},
pp.\ 609--616. Cambridge, MA: MIT Press.

[2] Bower, J.M.\ \& Beeman, D.\ (1995) {\it The Book of GENESIS: Exploring
  Realistic Neural Models with the GEneral NEural SImulation System.}  New York:
TELOS/Springer--Verlag.

[3] Hasselmo, M.E., Schnell, E.\ \& Barkai, E.\ (1995) Dynamics of learning and
recall at excitatory recurrent synapses and cholinergic modulation in rat
hippocampal region CA3. {\it Journal of Neuroscience} {\bf 15}(7):5249-5262.
\fi
\end{document}